\documentclass{mva_style}
\usepackage{graphicx}
\usepackage{subcaption}
\usepackage{algorithm}
\usepackage{algorithmic}
\usepackage{amssymb}
\usepackage{amsmath}
\usepackage{color}
\usepackage{booktabs}
\usepackage{hyperref}

\finalcopy 

\begin{document}
\title{FlowLoss: Dynamic Flow-Conditioned Loss Strategy for \\ Video Diffusion Models}

\author{
  \bf{Kuanting Wu}\thanks{\tt torridfish@elsa.cs.nthu.edu.tw} \\
  Science Tokyo \\
  \and
  \bf{Kei Ota}\thanks{\tt Ota.Kei@ds.MitsubishiElectric.co.jp}\\
  Mitsubishi Electric \\
  \and
  \bf{Asako Kanezaki}\thanks{\tt kanezaki@comp.isct.ac.jp} \\
  Science Tokyo \\
}

\maketitle

\section*{\centering Abstract}
\textit{
Video Diffusion Models (VDMs) can generate high-quality videos, but often struggle with producing temporally coherent motion. Optical flow supervision is a promising approach to address this, with prior works commonly employing warping-based strategies that avoid explicit flow matching. In this work, we explore an alternative formulation, FlowLoss, which directly compares flow fields extracted from generated and ground-truth videos. To account for the unreliability of flow estimation under high-noise conditions in diffusion, we propose a noise-aware weighting scheme that modulates the flow loss across denoising steps. Experiments on robotic video datasets suggest that FlowLoss improves motion stability and accelerates convergence in early training stages. Our findings offer practical insights for incorporating motion-based supervision into noise-conditioned generative models.
}

\section{Introduction}

Video Diffusion Models (VDMs)~\cite{ho2022video} have demonstrated impressive capabilities in synthesizing high-quality videos across diverse styles and domains~\cite{blattmann2023align, skorokhodov2024hierarchical, esser2023structure, xu2024magicanimate, wang2023modelscope}. Nevertheless, while these models can generate visually realistic outputs, they often lack physical consistency, which limits their applicability in downstream tasks such as robotic manipulation or physics-aware video prediction.

One possible remedy is to introduce optical flow as a supervision signal, which captures pixel-level motion dynamics. Prior methods have incorporated flow either as input conditioning~\cite{mathis2024onlyflow, shi2024motioni2v, hyelin2025opticalflow, jin2025flovd} or static auxiliary loss~\cite{feng2023flowvid, ni2023conditional, nam2024optical, yu2020learning}, but typically ignore the varying reliability of flow under different noise levels during diffusion. As a result, flow supervision can become unstable or ineffective---especially in high-noise training stages.

\begin{figure}[t]
\centering
    \includegraphics[width=\linewidth]{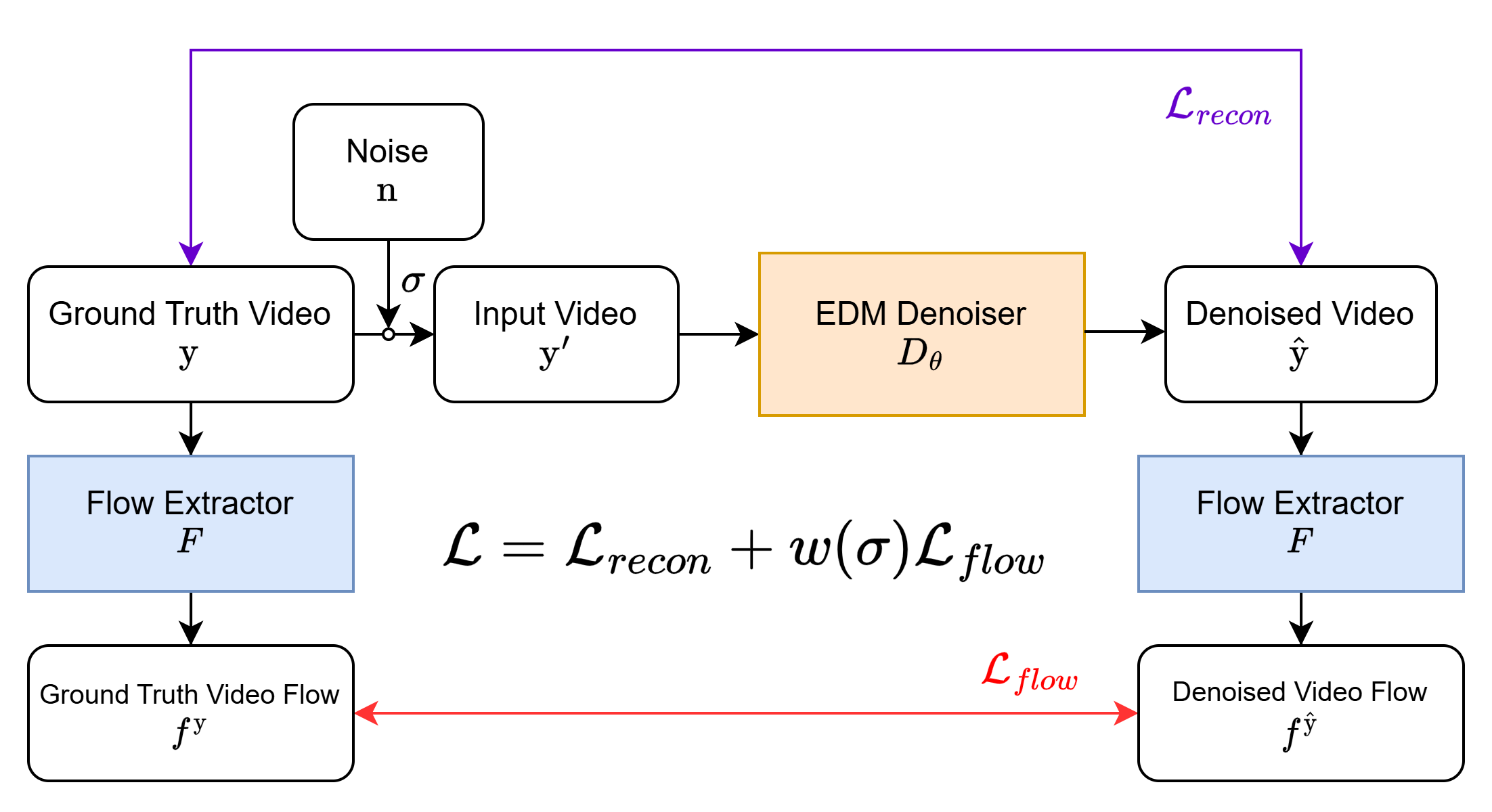}
    \caption{Overview of the FlowLoss supervision framework. The model is supervised by two gradient flows---one from pixel-level reconstruction $\mathcal{L}_\text{recon}$ and one from optical flow consistency $\mathcal{L}_\text{flow}$, which is adjusted by a dynamic weighting $w(\sigma)$, enabling it to generate visually plausible videos with coherent motion dynamics.}
    \label{fig:flowloss_arch}
\end{figure}

In this work, we propose \textbf{FlowLoss}, a noise-aware flow-conditioned loss strategy for VDMs. We introduce a differentiable flow loss that compares motion fields between generated and real videos, and modulates its contribution dynamically based on the noise scale $\sigma$. This design leverages reliable flow signals in low-noise regimes while avoiding unstable gradients in noisy inputs. Unlike prior works~\cite{feng2023flowvid, ni2023conditional, nam2024optical, yu2020learning} that rely on warping-based objectives, our method infers optical flow directly from video pairs using a differentiable flow extractor. As shown in Figure~\ref{fig:flowloss_arch}, our design allows gradients to propagate through both spatial and temporal dimensions, enabling end-to-end training of motion consistency without assuming flow correctness at the pixel level. Experiments on robotic video datasets show that FlowLoss accelerates convergence in early training stages. This offers practical insights for incorporating motion-based supervision into noise-conditioned generative models and suggests a promising direction for future research on accelerating convergence.

\begin{figure*}[h]
    \centering
    \begin{subfigure}[b]{0.24\textwidth}
        \includegraphics[width=\linewidth]{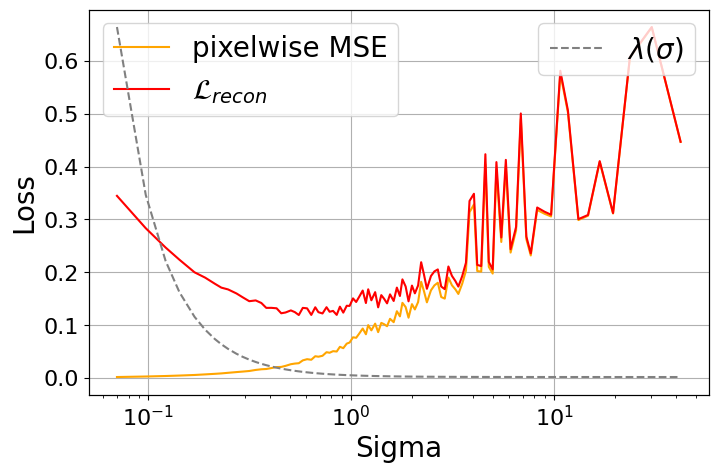}
        \caption{EDM loss function}
        \label{fig:vis_loss:sub1}
    \end{subfigure}
    \begin{subfigure}[b]{0.24\textwidth}
        \includegraphics[width=\linewidth]{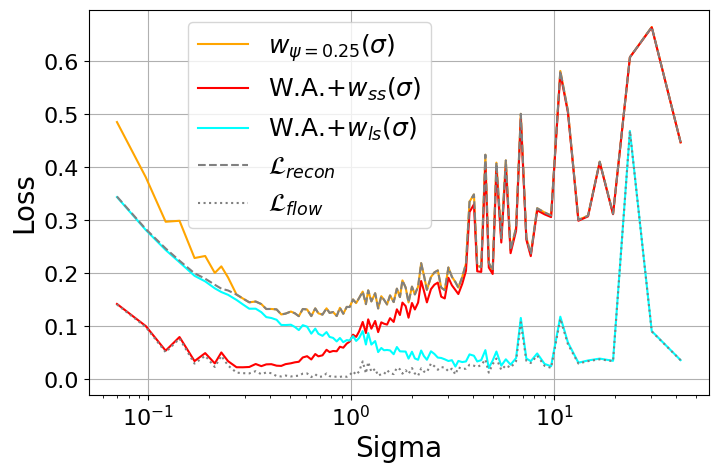}
        \caption{Our loss function}
        \label{fig:vis_loss:sub2}
    \end{subfigure}
    \begin{subfigure}[b]{0.24\textwidth}
        \includegraphics[width=\linewidth]{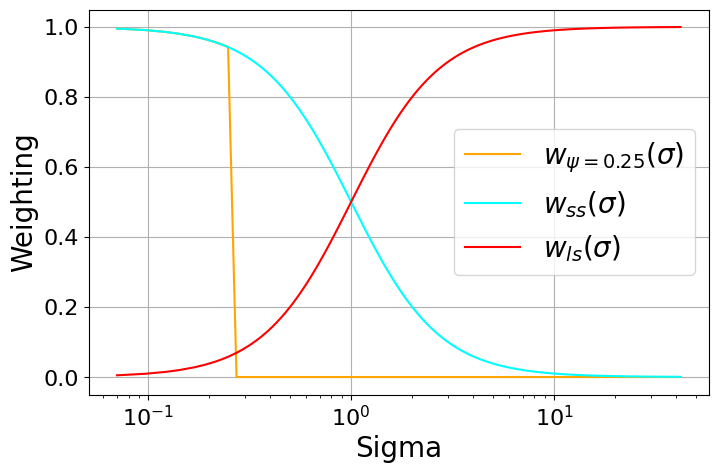}
        \caption{Weighting stratgies}
        \label{fig:vis_loss:sub3}
    \end{subfigure}
    \begin{subfigure}[b]{0.24\textwidth}
        \includegraphics[width=\linewidth]{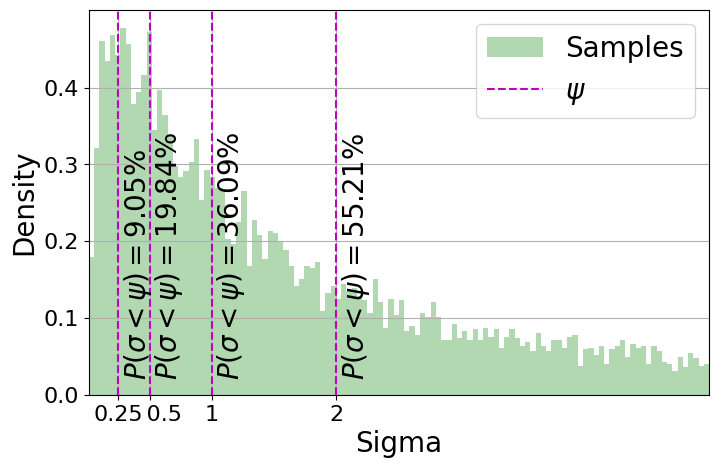}
        \caption{$\psi$ decision}
        \label{fig:vis_loss:sub4}
    \end{subfigure}
    \caption{
        Panels (a) and (b) show $\mathcal{L}_{\text{recon}}$ and $\mathcal{L}_{\text{flow}}$ computed on a single validation sample, using a VDM built upon the UNet backbone from \href{https://huggingface.co/stabilityai/stable-video-diffusion-img2vid}{Stable Video Diffusion 2.1 (Image-to-Video)}.
        \textbf{(a)} The original EDM defines the reconstruction loss as $\mathcal{L}_{\text{recon}} = \lambda(\sigma) \cdot \mathcal{L}_{\text{MSE}}$, where $\lambda(\sigma)$ increases as $\sigma$ decreases, encouraging fine-detail reconstruction during low-noise steps.  
        \textbf{(b)} Variants of our loss function. 
        \textbf{(c)} Corresponding weighting strategies. The $w_\psi(\sigma)$ function clips off $\mathcal{L}_{\text{flow}}$ contributions entirely when $\sigma$ exceeds a threshold, balancing cost and supervision strength.  
        \textbf{(d)} Distribution of sampled $\sigma$ values using EDM’s noise prior. Dashed lines show $\psi$ thresholds; percentages indicate the portion of steps where $\mathcal{L}_{\text{flow}}$ is applied under $w_\psi(\sigma)$.
    }
    \label{fig:flow_strategy}
\end{figure*}

\section{Related Work}

\noindent \textbf{Video Diffusion Models and Applications.}
Recent advances in diffusion-based generative models have led to significant progress in video synthesis. In parallel, numerous studies have explored improving the controllability of Video Diffusion Models (VDMs), enabling their application in a wider range of scenarios~\cite{shi2024motioni2v, zhang2023controlnet, gu2025diffusion, wu2024draganything, li2025image, geng2024motion, wang2024motionctrl, chen2025perception, guo2024sparsectrl}. As the field matures, VDMs have been increasingly adopted for downstream tasks such as robotics, where generated sequences serve as inputs for planning, control, or imitation~\cite{ko2023learning, wang2024language, du2023learning, yang2023unisim, soni2024videoagent}. These applications underscore the importance of generating videos with temporally consistent and physically plausible motion.

\noindent \textbf{Flow as Loss Supervision.}
Other approaches leverage optical flow as an auxiliary loss to enforce temporal consistency. Examples include FlowVid~\cite{feng2023flowvid}, latent flow diffusion models (LFDM)~\cite{ni2023conditional}, and temporal stabilizers such as~\cite{nam2024optical, yu2020learning}, which typically use pre-trained flow extractors (e.g. FlowNet~\cite{fischer2015flownet} or PCAFlow~\cite{wulff2015efficient}) in a differentiable but static way, often through image warping between frames to minimize pixel-level discrepancies.

Dense optical flow captures pixel-level motion dynamics, it serves as a valuable signal for enforcing temporal coherence. In contrast to prior work, our method leverages this by computing a differentiable flow loss through direct comparison of dense flow fields extracted from generated outputs and ground-truth videos.

\noindent \textbf{EDM-based Denoising.}
Our work builds upon the formulation introduced in the Diffusion-Based Generative Model Design Space (EDM)~\cite{karras2022elucidating}. Given a clean video $\text{y}$, a noisy input $\text{y}'$ is constructed by adding Gaussian noise $\text{n} \sim \mathcal{N}(0, 1)$ scaled by a sampled noise level $\sigma$, where $\ln(\sigma) \sim \mathcal{N}(P_{\text{mean}}, P_{\text{std}}^2)$. The denoised video $\hat{\text{y}}$ is generated by denoising model $D_\theta(\text{y}'; \sigma)$, which is trained by minimizing the following loss:
\[
\mathcal{L}_{\text{recon}} = \mathbb{E}_{\sigma, \text{y}, \text{n}}\left[\lambda(\sigma) \cdot \| D_\theta(\text{y} + \text{n}; \sigma) - y \|^2 \right],
\]
where $\lambda(\sigma)=\frac{\sigma^2+1}{\sigma^2}$ serves as a noise-aware weighting function to balance gradient magnitudes across different noise levels, as illustrated in Figure~\ref{fig:vis_loss:sub1}. In our method, we extend this formulation to incorporate a flow-based loss, leveraging the similar $\sigma$-centric design to enable dynamic supervision scheduling.

\section{Methodology}
\label{method}

\begin{figure}[t]
\centering
    \includegraphics[width=\linewidth]{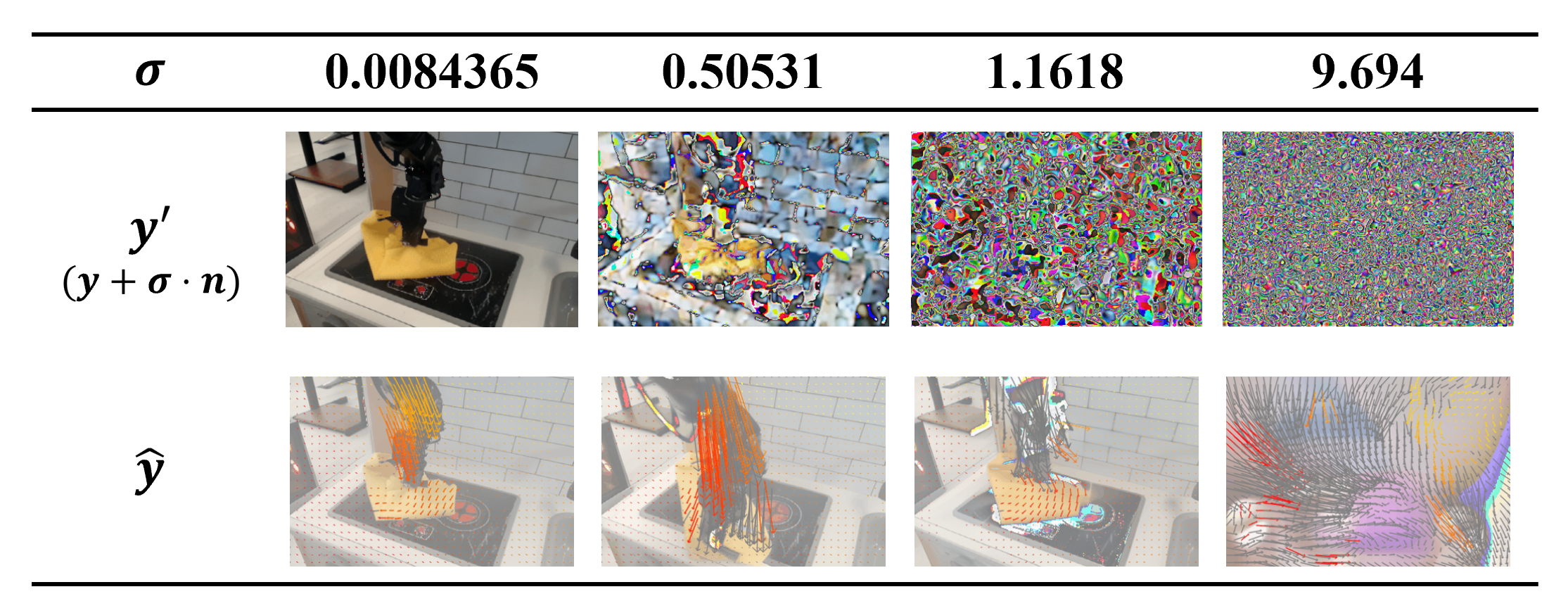}
\caption{Higher noise scales $\sigma$ lead to corrupted inputs and degraded flow extraction, motivating our noise-aware flow loss design.
}
\label{fig:noise}
\end{figure}

\begin{figure*}[t]
\centering
    \begin{subfigure}[b]{0.15\textwidth}
        \includegraphics[width=\linewidth]{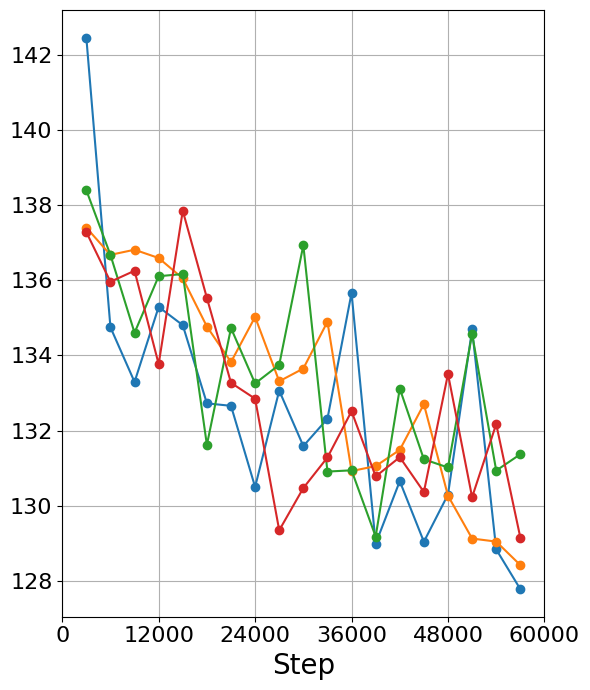}
        \caption{FID}
        \label{fig:validation-result:sub1}
    \end{subfigure}
    \begin{subfigure}[b]{0.15\textwidth}
        \includegraphics[width=\linewidth]{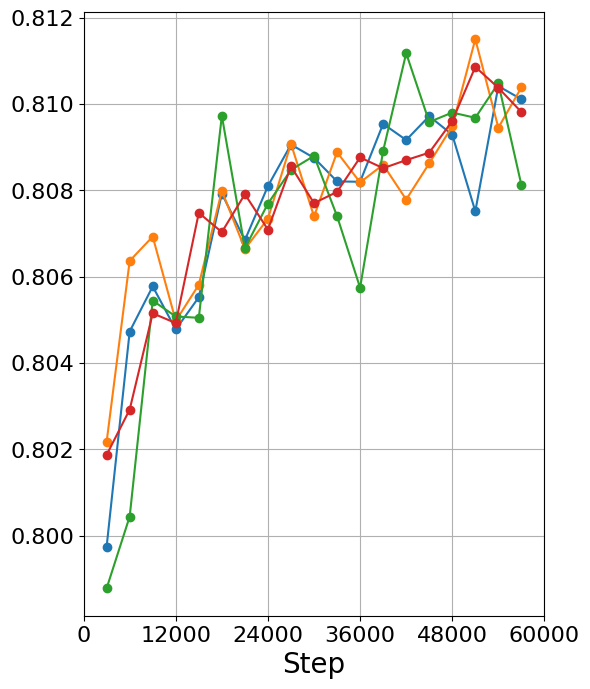}
        \caption{SSIM}
        \label{fig:validation-result:sub2}
    \end{subfigure}
    \begin{subfigure}[b]{0.15\textwidth}
        \includegraphics[width=\linewidth]{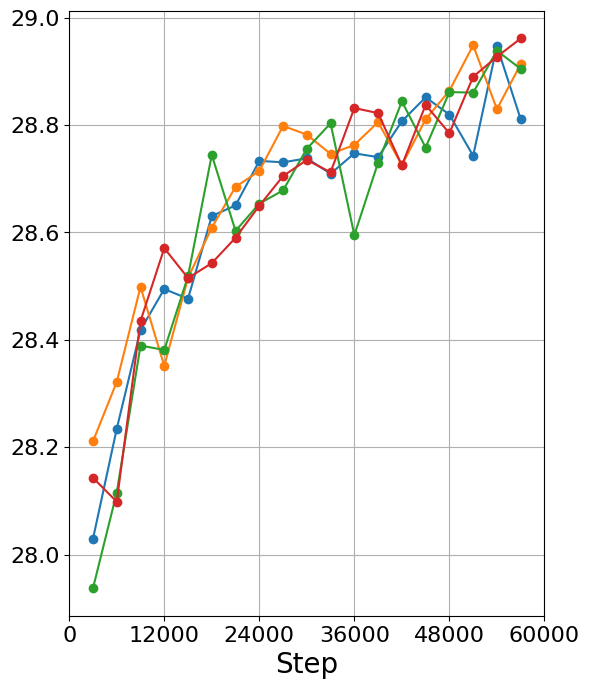}
        \caption{PSNR}
        \label{fig:validation-result:sub3}
    \end{subfigure}
    \begin{subfigure}[b]{0.15\textwidth}
        \includegraphics[width=\linewidth]{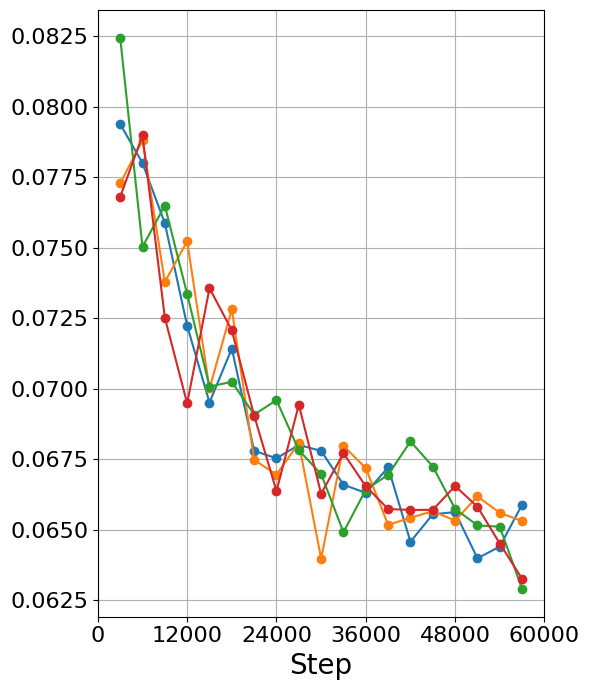}
        \caption{LPIPS}
        \label{fig:validation-result:sub4}
    \end{subfigure}
    \begin{subfigure}[b]{0.15\textwidth}
        \includegraphics[width=\linewidth]{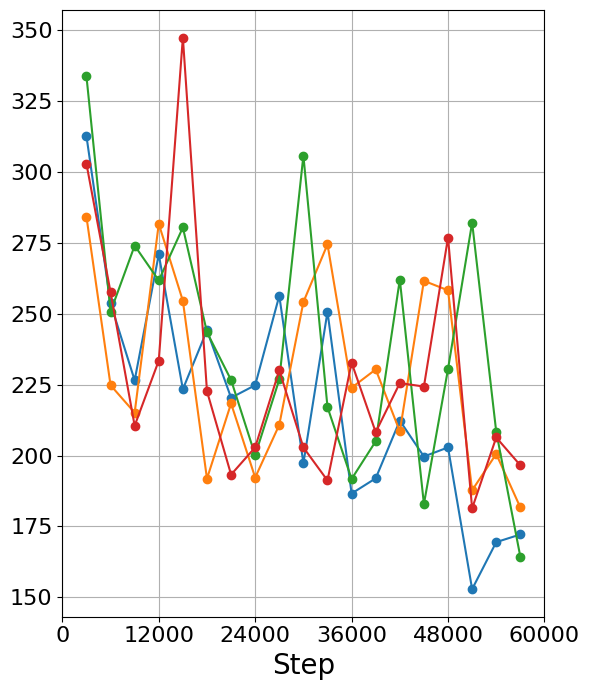}
        \caption{FVD}
        \label{fig:validation-result:sub5}
    \end{subfigure}
    \begin{subfigure}[b]{0.15\textwidth}
        \includegraphics[width=\linewidth]{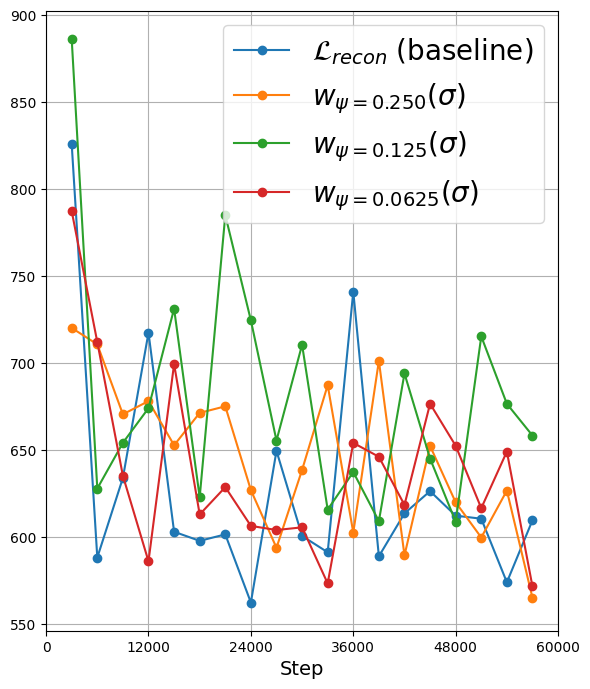}
        \caption{$\mathcal{L}_{\text{flow}}/\lambda(\sigma)$}
        \label{fig:validation-result:sub6}
    \end{subfigure}
  \caption{Validation performance over training steps for different flow loss strategies.}
  \label{fig:validation-result}
\end{figure*}

To improve temporal consistency and physical plausibility in video generation, we introduce a training objective that incorporates optical flow supervision directly.
In this section, we first define the flow-based loss term $\mathcal{L}_{\text{flow}}$, then describe how it is integrated with the EDM-style reconstruction loss $\mathcal{L}_{\text{recon}}$, and finally present a noise-aware gating strategy designed to balance flow supervision with computational efficiency.

\noindent \textbf{Flow Loss.}
We use a pretrained dense flow estimator $F$ (DOT~\cite{le2024dense}) to extract pixel-wise flow from both ground truth $y$ and the denoised prediction $\hat{y}$, yielding $f^y$ and $f^{\hat{y}}$. For each pixel at location $(i, j)$, the flow field $f_t(i, j) \in \mathbb{R}^2$ represents its motion vector at time $t$, while the binary occlusion mask $\alpha_t(i, j) \in \{0, 1\}$ indicates whether the pixel is visible.

The flow loss is defined as:
\[
\mathcal{L}_{\text{flow}} = s \cdot \mathbb{E}_{\sigma, y, n} \left[ \lambda(\sigma) \sum_{t} o(\alpha^{\text{y}}_t) \cdot \| f_t^y - f_t^{\hat{y}} \|^2_2 \right],
\]
where $o(\alpha_t(i,j)) = 1$ for visible regions and $0.3$ otherwise. The term $\lambda(\sigma) = \frac{\sigma^2 + 1}{\sigma^2}$ follows the EDM weighting scheme, and $s=10^{-6}$ is a global scaling factor to balance factor loss and reconstruction loss. This loss emphasizes motion consistency in visible areas, while reducing sensitivity to unreliable or occluded regions.

\noindent \textbf{Full Training Objective.}
We combine the standard reconstruction loss $\mathcal{L}_{\text{recon}}$ with flow supervision:
\[
\mathcal{L} = \mathcal{L}_{\text{recon}} + w_{\psi}(\sigma) \cdot \mathcal{L}_{\text{flow}},
\]
where $w_\psi(\sigma)$ scales the flow loss based on the noise level $\sigma$. Since flow predictions degrade with increasing noise (see Figure~\ref{fig:noise}), this dynamic weighting helps avoid introducing harmful gradients at high-noise steps.

\noindent \textbf{Hard Gating Strategy.}
To reduce computation, we further adopt a hard gating mechanism:
\[
w_{\psi}(\sigma)=\begin{cases}
\frac{1}{\sigma^2+1} & \text{if } \sigma < \psi \\
0 & \text{otherwise}
\end{cases}
\]
This ensures $\mathcal{L}_{\text{flow}}$ is only applied when flow supervision is likely reliable. As EDM samples $\sigma$ from a long-tailed log-normal distribution, many steps fall into high-noise regimes. Our gating scheme allows computation to focus on cleaner inputs, where flow guidance is most effective (Figure~\ref{fig:vis_loss:sub3}~\ref{fig:vis_loss:sub4}).

With our training objective defined, we proceed to validate its impact on video generation quality and motion consistency.

\section{Experiments}

\setlength{\textfloatsep}{6pt}
\begin{table}[t]
  \caption{Quantitative result of comparing our method with baseline. All of them are trained with 57,000 steps and is evaluated on test dataset. Bold font denotes the best result.}
  \begin{center}
    \resizebox{\linewidth}{!}{%
      \begin{tabular}{@{}c c c c c c c@{}}
        \toprule
        Method & FID $\downarrow$ & SSIM $\uparrow$ & PSNR $\uparrow$ & LPIPS $\downarrow$ & FVD $\downarrow$ &  Training time \\
        \midrule
        baseline  ($\mathcal{L} = \mathcal{L}_{\text{recon}}$) & \textbf{127.6907} & \textbf{0.8118} & 28.8986 & 0.0651 & 165.3577 & 07h 56m 47s \\
        ours + $w_{\psi=0.0625}(\sigma)$ & 129.1449 & 0.8112 & \textbf{29.0284} & 0.0626 & 191.2058 &  08h 45m 08s \\
        ours + $w_{\psi=0.125}(\sigma)$ & 130.1602  & 0.8097 &  28.9757 &  \textbf{0.0625} & \textbf{161.3095} & 10h 31m 28s \\
        ours + $w_{\psi=0.250}(\sigma)$ & 128.6550  & 0.8118 &  28.9896 & 0.0649 & 179.0681 & 14h 29m 26s \\
        \bottomrule
      \end{tabular}
    }
    \label{tab:psi-comparison}
  \end{center}
\end{table}

\begin{figure}[t]
\centering
    \includegraphics[width=\linewidth]{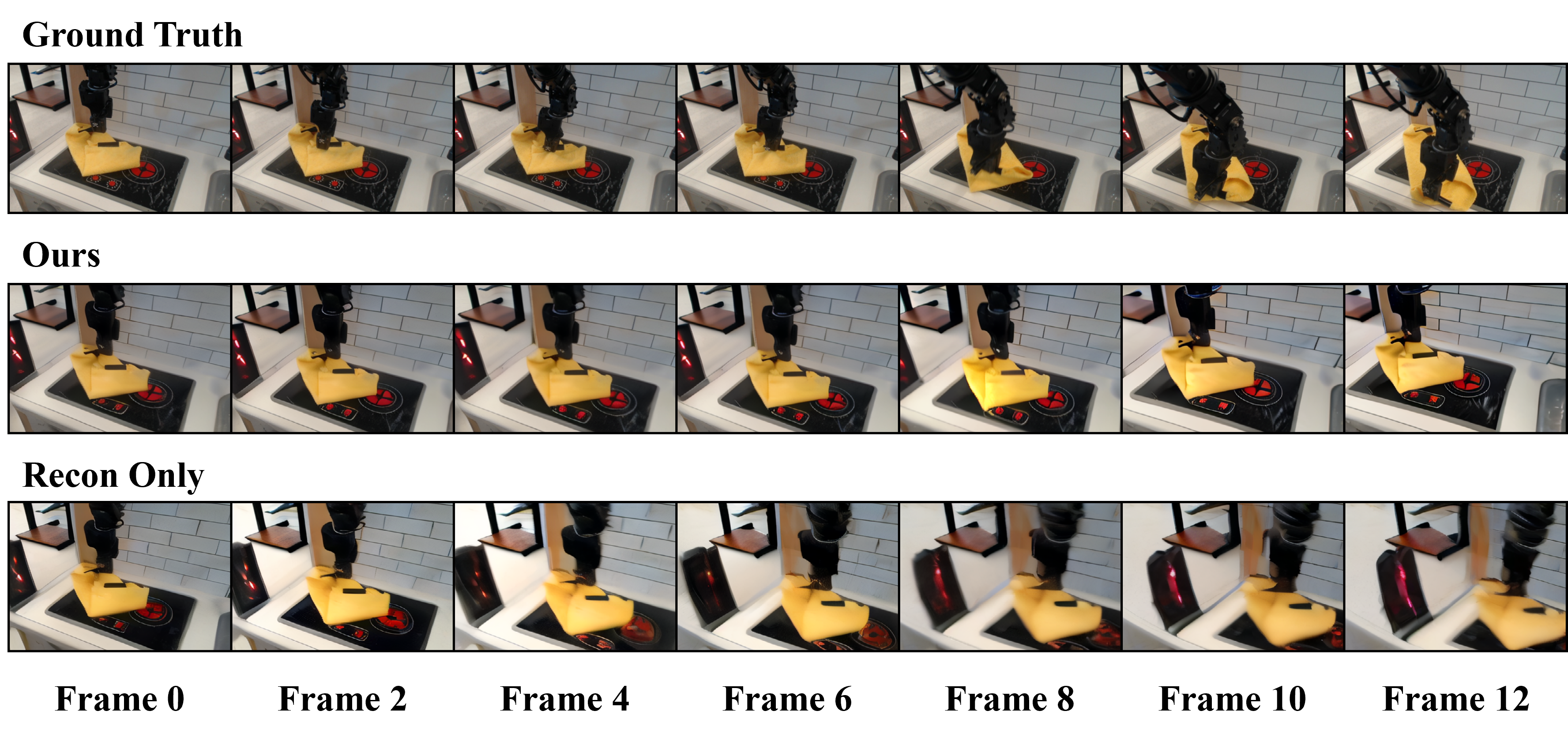}
\caption{Comparison of early-stage generation results (step = 100) across different training objectives.
Ground Truth refers to the original robot video sequence. Ours uses the full loss function $\mathcal{L}_{\text{recon}} + w(\sigma) \cdot \mathcal{L}_{\text{flow}}$, while Recon Only trains with reconstruction loss $\mathcal{L}_{\text{recon}}$ alone. Our method exhibits improved motion stability and temporal coherence at an early diffusion step, whereas the baseline suffers from spatial drift and jitter.}
\label{fig:early-stage}
\end{figure}

We evaluate our proposed loss designs on a robotic video dataset, comparing against reconstruction-only baselines.

\noindent \textbf{Experimental Setup.}
We build our VDM on the UNet backbone from Stable Video Diffusion 2.1 (Image-to-Video)\cite{blattmann2023stable}, using the updated implementation from\cite{wang2024language}. Our goal is to evaluate whether motion-guided supervision enhances temporal coherence—especially important in robotics, where downstream tasks rely on stable dynamics.

Experiments are conducted on the Bridge Dataset v2~\cite{walke2023bridgedata}, a large-scale collection of real-world robotic manipulation videos. It contains 12,850 training, 1,838 validation, and 3,672 test samples, each consisting of a natural language task prompt and a trajectory of 14 resampled frames at $384 \times 256$ resolution.

We adopt an \textbf{Image+Text-to-Video} setup, conditioning the model on the initial frame $I_0$ and task prompt $P$ via cross-attention and FiLM layers~\cite{perez2018film}, following the architecture from~\cite{wang2024language}.

Models are trained on a single NVIDIA H100 GPU for 57,000 steps. Noise levels $\sigma$ follow a log-normal sampling scheme from EDM~\cite{karras2022elucidating}. The reconstruction loss $\mathcal{L}{\text{recon}}$ is applied at each step, while the flow loss $\mathcal{L}{\text{flow}}$ is gated by a $\psi$-based schedule.

\noindent \textbf{Evaluation Metrics.}
We assess performance using both image-level and video-level metrics. For frame-level fidelity, we report FID~\cite{heusel2017gans}, SSIM~\cite{zhou2004ssim}, PSNR~\cite{hore2010psnr}, and LPIPS~\cite{zhang2018unreasonable}. For temporal consistency, we report FVD~\cite{unterthiner2018towards} and unweighted $\mathcal{L}_{\text{flow}}$. All evaluations are conducted on held-out validation sequences with consistent frame counts and prompt formats.

\noindent \textbf{Early-stage stabilization.}
Standard VDMs often produce unstable outputs early in training, especially on fixed-view datasets like Bridge v2, where most pixels are static. This instability is especially pronounced during early denoising steps. In contrast, our method yields more stable outputs as early as step 100 (Figure~\ref{fig:early-stage}), suggesting it helps the model quickly capture the global scene structure and adapt to its static layout.

\noindent \textbf{Outcome analysis.}
While early-stage motion stabilization is evident, the final quantitative results paint a more nuanced picture. Although Table~\ref{tab:psi-comparison} indicates that our method slightly outperforms the baseline across several metrics, Figure~\ref{fig:validation-result} reveals that both methods follow similar trends during training, with no consistent advantage at convergence. In some cases, our method shows marginal improvements; in others, it performs slightly worse. We attribute this outcome to two primary factors: (1) the hard gating mechanism $w_{\psi}(\sigma)$ may excessively suppress flow supervision at high noise levels, limiting its overall influence on learning; and (2) the flow extractor (e.g., DOT~\cite{le2024dense}) may produce noisy or inaccurate gradients—especially in occluded or low-texture regions—reducing the effectiveness of the supervision signal.

In spite of that, the early flattening of validation curves across FVD and $\mathcal{L}_{\text{flow}}$ suggests that flow supervision—when active—helps the model reach a reasonable motion prior more quickly. This property could be beneficial for accelerating VDM training or guiding curriculum-style optimization schedules. Despite the lack of significant final-stage improvements, our results indicate that flow-based guidance has the potential to enhance sample efficiency and training stability, particularly during the most uncertain phases of denoising.

\section{Ablation Study}

\begin{figure}[t]
\centering
    \includegraphics[width=\linewidth]{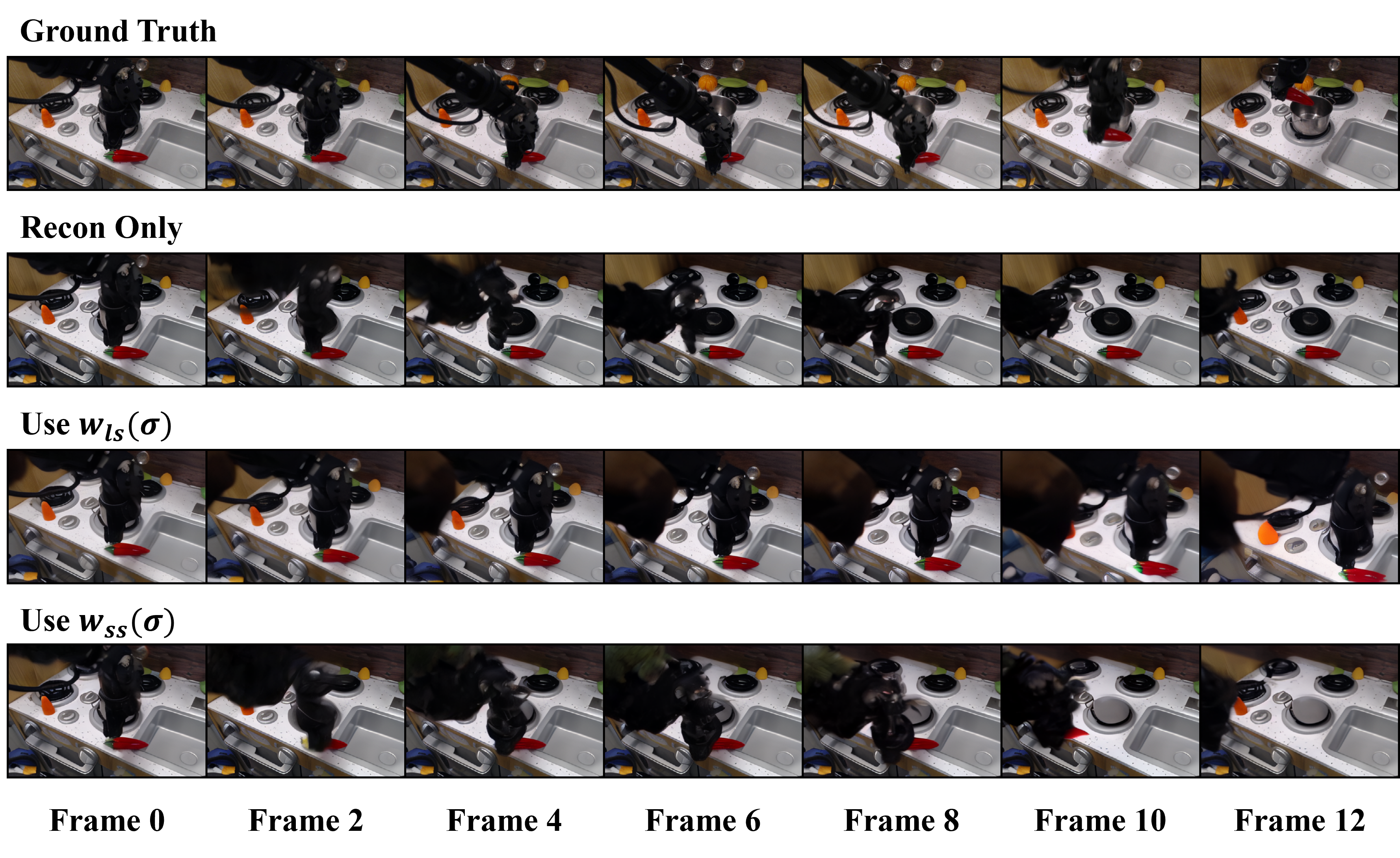}
\caption{Comparison of generated videos at 1,600 training steps using different weighting strategies under the weighted average formulation.}
\label{fig:sigma-focus}
\end{figure}

In addition to our main approach, we explore an alternative method for combining $\mathcal{L}_{\text{recon}}$ and $\mathcal{L}_{\text{flow}}$, which we refer to as the \textbf{Weighted Average} strategy:

\[
\mathcal{L} = (1 - w(\sigma)) \cdot \mathcal{L}_{\text{recon}} + w(\sigma) \cdot \mathcal{L}_{\text{flow}},
\]

Unlike our default formulation, which adds flow loss on top of reconstruction loss, this design emphasizes one loss exclusively depending on the noise level $\sigma$. As shown in Figure~\ref{fig:vis_loss:sub2}, when $\sigma$ is small or large, only one of the two losses is active while the other is completely suppressed. To test the effectiveness of this strategy, we design two contrasting weighting schedules:
\begin{itemize}
\item $w_{\text{ss}}(\sigma)$: Emphasizes flow supervision in small-$\sigma$ (clean input) regimes.
\item $w_{\text{ls}}(\sigma)$: Emphasizes flow supervision in large-$\sigma$ (noisy input) regimes.
\end{itemize}
These variants serve to validate our hypothesis that flow supervision is more effective when applied in low-noise stages, where flow extraction is more reliable. As shown in Figure~\ref{fig:sigma-focus}, after only 1,600 training steps, both the baseline model (with $\mathcal{L}_{\text{recon}}$ only) and the $w_{\text{ss}}(\sigma)$ variant achieve stable results with minimal jitter. In contrast, the $w_{\text{ls}}(\sigma)$ variant continues to produce unstable, flickering outputs.

These findings suggest that applying flow loss in high-noise regimes may introduce harmful supervision signals due to noisy or inaccurate flow estimates. This ablation further reinforces the importance of noise-aware scheduling and supports our core intuition: flow guidance should be selectively applied only when motion signals are reliable.

\section{Conclusion and Discussion}

In this work, we introduced FlowLoss, a noise-aware flow-conditioned loss strategy for Video Diffusion Models (VDMs). By dynamically adjusting the contribution of flow loss based on noise levels, our approach addresses the challenge of maintaining temporal consistency in generated videos. Our experiments on robotic video datasets show that FlowLoss enables faster convergence in early training stages and improves motion stability. However, it also introduces significant computational overhead.

Our findings highlight both the potential and limitations of flow-based supervision in VDMs. While the dynamic weighting strategy mitigates the impact of noise on flow consistency, further improvements may be achievable by refining the gating mechanism and exploring more robust flow extractors. This work lays the groundwork for future research into incorporating motion-aware losses in noise-conditioned generative models.

\bibliographystyle{unsrt}
\bibliography{reference}
\end{document}